\documentclass[10pt,twocolumn,letterpaper]{article}

\usepackage{iccv}
\usepackage{times}
\usepackage{epsfig}
\usepackage{graphicx}
\usepackage{amsmath}
\usepackage{amssymb}
\usepackage{xcolor}
\usepackage{tabu}
\usepackage{widetext}
\usepackage{booktabs}
\usepackage{multirow}
\usepackage{algorithm}
\usepackage{algorithmic}
\usepackage[utf8]{inputenc}

\usepackage[pagebackref=true,breaklinks=true,letterpaper=true,colorlinks,bookmarks=false]{hyperref}

\iccvfinalcopy 

\def\iccvPaperID{1656} 

\newcommand{\tabincell}[2]{\begin{tabular}{@{}#1@{}}#2\end{tabular}}  

\providecommand{\tabularnewline}{\\}

\begin{document}


\title{Pixel2Mesh++: Multi-View 3D Mesh Generation via Deformation}


\author{Chao Wen$^{1}$\thanks{indicates equal contributions.} \qquad Yinda Zhang$^{2}$\footnotemark[1] \qquad Zhuwen Li$^{3}$\footnotemark[1] \qquad Yanwei Fu$^{1}$\thanks{indicates corresponding author. This work is supported by the STCSM project (19ZR1471800), and Eastern Scholar (TP2017006). }\\
$^1$Fudan University \qquad $^2$Google LLC \qquad $^3$Nuro, Inc.}

\maketitle

\begin{abstract}
   
We study the problem of shape generation in 3D mesh representation from a few color images with known camera poses. 
While many previous works learn to hallucinate the shape directly from priors, we resort to further improving the shape quality by leveraging cross-view information with a graph convolutional network.
Instead of building a direct mapping function from images to 3D shape, our model learns to predict series of deformations to improve a coarse shape iteratively.
Inspired by traditional multiple view geometry methods, our network samples nearby area around the initial mesh's vertex locations and reasons an optimal deformation using perceptual feature statistics built from multiple input images.
Extensive experiments show that our model produces accurate 3D shape that are not only visually plausible from the input perspectives, but also well aligned to arbitrary viewpoints.
With the help of physically driven architecture, our model also exhibits generalization capability across different semantic categories, number of input images, and quality of mesh initialization.
   
\end{abstract}

\section{Introduction}
3D shape generation has become a popular research topic recently. 
With the astonishing capability of deep learning, lots of works have been demonstrated to successfully generate the 3D shape from merely a single color image.
However, due to limited visual evidence from only one viewpoint, single image based approaches usually produce rough geometry in the occluded area and do not perform well when generalized to test cases from domains other than training, e.g. cross semantic categories.

\begin{figure}[t]
	\centering
    	\includegraphics[width=\columnwidth]{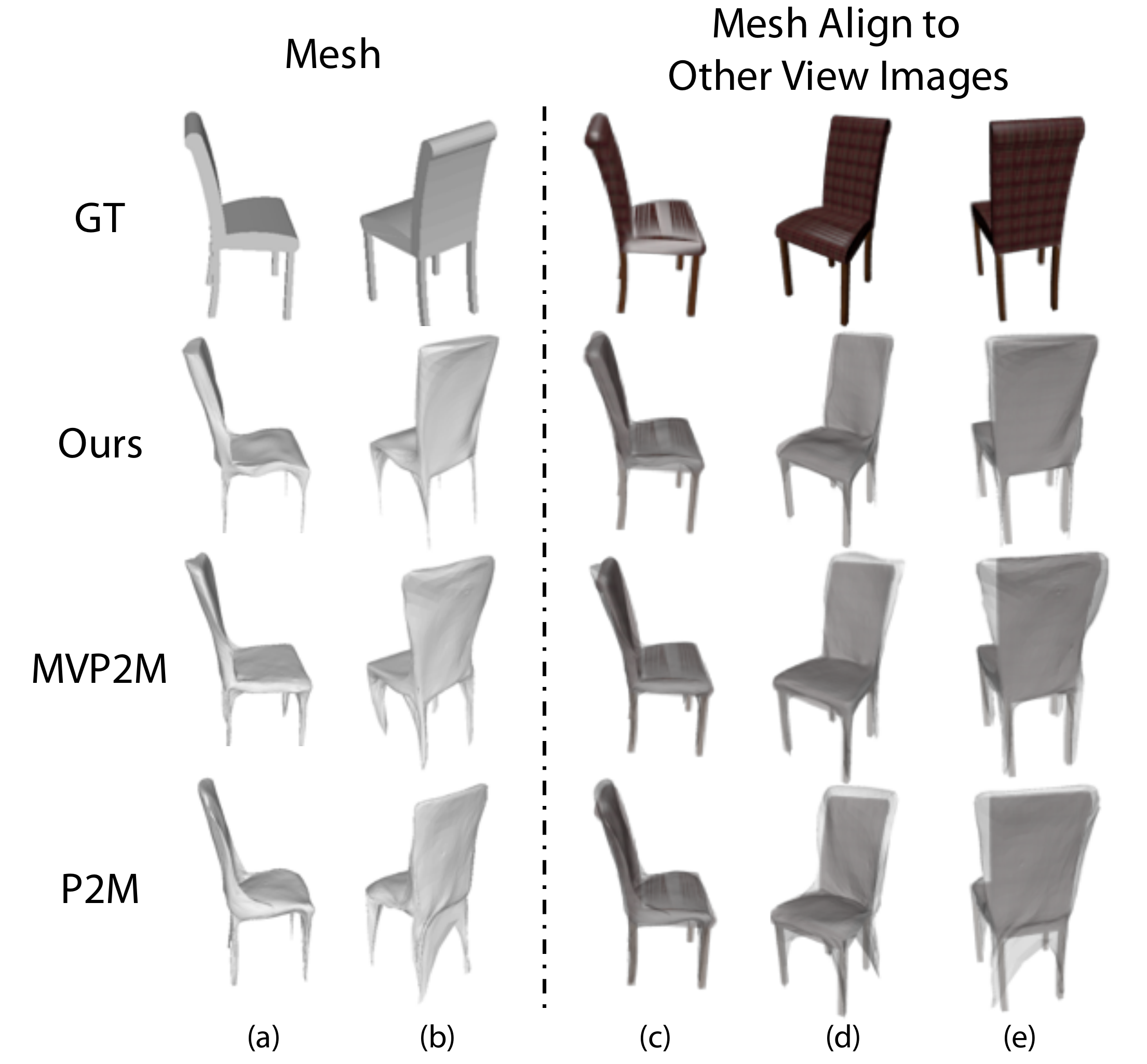}
	\caption{{\bf Multi-View Shape Generation.} From multiple input images, we produce shapes aligning well to input (c and d) and arbitrary random (e) camera viewpoint.
	Single view based approach, e.g. Pixel2Mesh (P2M) \cite{wang2018pixel2mesh}, usually generates shape looking good from the input viewpoint (c) but significantly worse from others.
	Naive extension with multiple views (MVP2M, Sec. \ref{sec:mvp2m}) does not effectively improve the quality.}
	\label{fig:teaser}
 	\vspace{-3mm}
\end{figure}

Adding a few more images (e.g. 3-5) of the object is an effective way to provide the shape generation system with more information about the 3D shape.
On one hand, multi-view images provide more visual appearance information, and thus grant the system with more chance to build the connection between 3D shape and image priors.
On the other hand, it is well-known that traditional multi-view geometry methods \cite{HarlteyZ2001} accurately infer 3D shape from correspondences across views, which is analytically well defined and less vulnerable to the generalization problem. 
However, these methods typically suffer from other problems, like large baselines and poorly textured regions.
Though typical multi-view methods are likely to break down with very limited input images (e.g. less than 5), the cross-view connections might be implicitly encoded and learned by a deep model. 
While well-motivated, there are very few works in the literature exploiting in this direction, and a naive multi-view extension of single image based model does not work well as shown in Fig. \ref{fig:teaser}.

In this work, we propose a deep learning model to generate the object shape from multiple color images.
Especially, we focus on endowing the deep model with the capacity of improving shapes using cross-view information.
We resort to designing a new network architecture, named Multi-View Deformation Network (MDN), which works in conjunction with the Graph Convolutional Network (GCN) architecture proposed in Pixel2Mesh \cite{wang2018pixel2mesh} to generate accurate 3D geometry shape in the desirable mesh representation.
In Pixel2Mesh, a GCN is trained to deform an initial shape to the target using features from a single image, which often produces plausible shapes but lack of accuracy (Fig. \ref{fig:teaser} P2M).
We inherit this characteristic of ``generation via deformation'' and further deform the mesh in MDN using features carefully pooled from multiple images.
Instead of learning to hallucinate via shape priors like in Pixel2Mesh, MDN reasons shapes according to correlations across different views through a physically driven architecture inspired by classic multi-view geometry methods.
In particular, MDN proposes hypothesis deformations for each vertex and move it to the optimal location that best explains features pooled from multiple views.
By imitating correspondences search rather than learning priors, MDN generalizes well in various aspects, such as cross semantic category, number of input views, and the mesh initialization.

\begin{figure*}[h]
	\centering
	\includegraphics[width=0.98\textwidth]{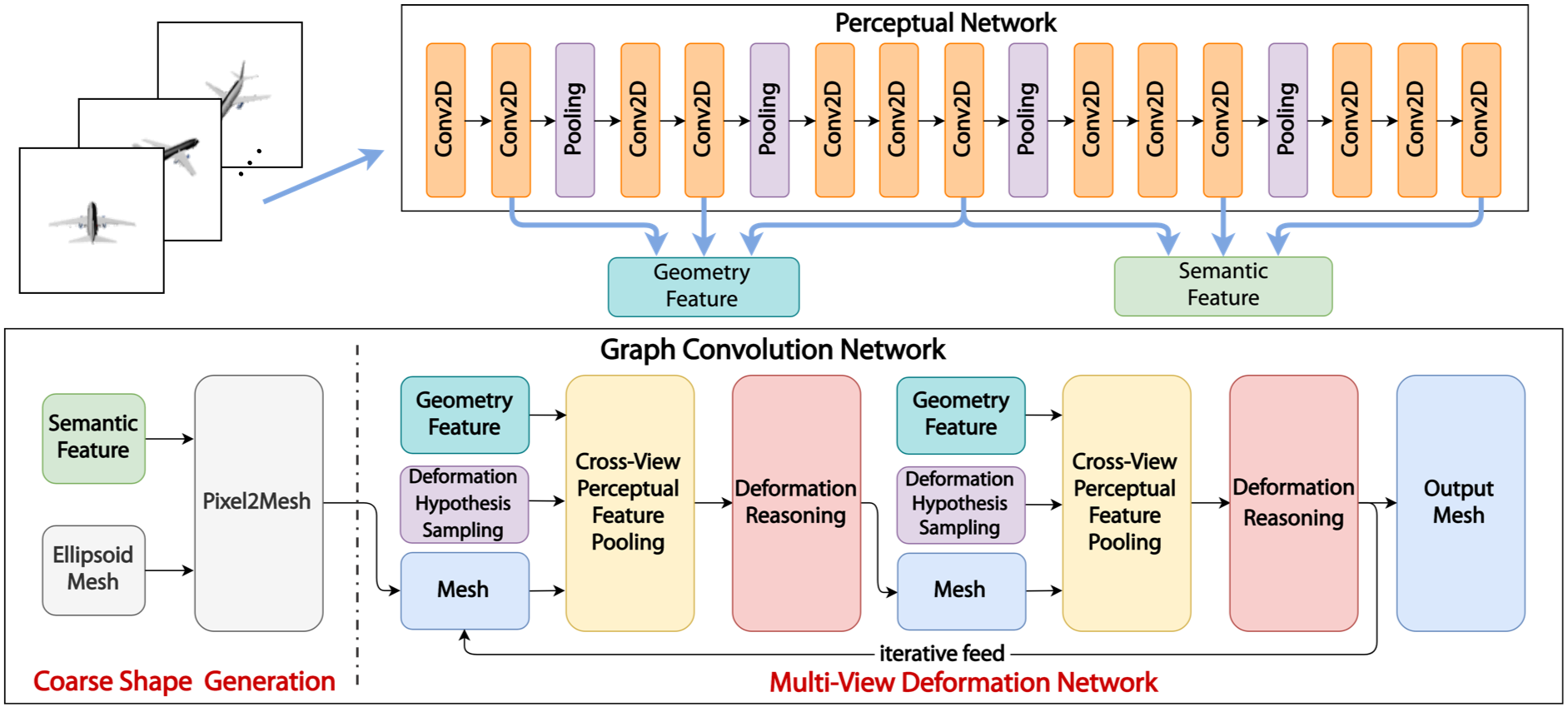}
	\caption{{\bf System Pipeline.} Our whole system consists of a 2D CNN extracting image features and a GCN deforming an ellipsoid to target shape. A coarse shape is generated from Pixel2Mesh and refined iteratively in Multi-View Deformation Network. To leverage cross-view information, our network pools perceptual features from multiple input images for hypothesis locations in the area around each vertex and predicts the optimal deformation.}
	\label{fig:pipeline} 
	\vspace{-2mm}
\end{figure*}

Besides the above-mentioned advantages, MDN is in addition featured with several good properties.
First, it can be trained end-to-end. Note that it is non-trivial since MDN searches deformation from hypotheses, which requires a non-differentiable argmax/min.
Inspired by \cite{KendallMDH17}, we apply a differentiable 3D soft argmax, which takes a weighted sum of the sampled hypotheses as the vertex deformation.
Second, it works with varying number of input views in a single forward pass. 
This requires the feature dimension to be invariant with the number of inputs, which is typically broken when aggregating features from multiple images (e.g. when using concatenation).
We achieve the input number invariance by concatenating the statistics (e.g. mean, max, and standard deviation) of the pooled feature, which further maintains input order invariance.
We find this statistics feature encoding explicitly provides the network cross-view information, and encourages it to automatically utilize image evidence when more are available.
Last but not least, the nature of ``generation via deformation'' allows an iterative refinement. 
In particular, the model output can be taken as the input, and quality of the 3D shape is gradually improved throughout iterations.
With these desiring features, our model achieves the state-of-the-art performance on ShapeNet for shape generation from multiple images under standard evaluation metrics.


To summarize, we propose a GCN framework that produces 3D shape in mesh representation from a few observations of the object in different viewpoints.
The core component is a physically driven architecture that searches optimal deformation to improve a coarse mesh using perceptual feature statistics built from multiple images, which produces accurate 3D shape and generalizes well across different semantic categories, numbers of input images, and the quality of coarse meshes.




\section{Related Work}

\paragraph{3D Shape Representations}
Since 3D CNN is readily applicable to 3D volumes, the volume representation has been well-exploited for 3D shape analysis and generation \cite{choy20163d,wang2017cnn}. With the debut of PointNet \cite{qi2016pointnet}, the point cloud representation has been adopted in many works \cite{fan2017point,QiLWSG18}. Most recently, the mesh representation \cite{KatoUH18,wang2018pixel2mesh} has become competitive due to its compactness and nice surface properties. Some other representations have been proposed, such as geometry images \cite{SinhaUHR17}, depth images \cite{TatarchenkoDB16,Richter018}, classification boundaries \cite{mescheder2019occupancy, chen2019learning}, signed distance function \cite{Park_2019_CVPR}, etc., and most of them require post-processing to get the final 3D shape. Consequently, the shape accuracy may vary and the inference take extra time.

\vspace{-3mm}
\paragraph{Single view shape generation}
Classic single view shape reasoning can be traced back to shape from shading \cite{DurouFS08,ZhangTCS99}, texture \cite{MarinosB90}, and de-focus \cite{FavaroS05}, which only reason the visible parts of objects. 
With deep learning, many works leverage the data prior to hallucinate the invisible parts, and directly produce shape in 3D volume \cite{choy20163d,GirdharFRG16,WuZXFT16,HaneTM17,RieglerUG17,TatarchenkoDB17,JohnstonGCR17}, point cloud \cite{fan2017point}, mesh models \cite{KatoUH18}, or as an assembling of shape primitive \cite{TulsianiSGEM17,Niu0018}.
Alternatively, 3D shape can be also generated by deforming an initialization, which is more related to our work.
Tulsiani \etal\cite{TulsianiKCM17} and Kanazawa \etal\cite{KanazawaTEM18} learn a category-specific 3D deformable model and reasons the shape deformations in different images. 
Wang \etal \cite{wang2018pixel2mesh} learn to deform an initial ellipsoid to the desired shape in a coarse to fine fashion.
Combining deformation and assembly, Huang \etal\cite{HuangWK15} and Su \etal\cite{SuHMKG14} retrieve shape components from a large dataset and deform the assembled shape to fit the observed image. 
Kuryenkov \etal \cite{kurenkov2018deformnet} learns free-form deformations to refine shape.
Even though with impressive success, most deep models adopt an encoder-decoder framework, and it is arguable if they perform shape generation or shape retrieval \cite{TatarchenkoRRLKB19}.

\vspace{-3mm}
\paragraph{Multi-view shape generation}
Recovering 3D geometry from multiple views has been well studied. 
Traditional multi-view stereo (MVS) \cite{HarlteyZ2001} relies on correspondences built via photo-consistency and thus it is vulnerable to large baselines, occlusions, and texture-less regions. 
Most recently, deep learning based MVS models have drawn attention, and most of these approaches \cite{yao2018mvsnet, huang2018deepmvs, im2018dpsnet, zhang2018activestereonet} rely on a cost volume built from depth hypotheses or plane sweeps. 
However, these approaches usually generate depth maps, and it is non-trivial to fuse a full 3D shape from them. 
On the other hand, direct multi-view shape generation uses fewer input views with large baselines, which is more challenging and has been less addressed. 
Choy \etal\cite{choy20163d} propose a unified framework for single and multi-view object generation reading images sequentially. Kar \etal\cite{kar2017learning} learn a multi-view stereo machine via recurrent feature fusion. Gwak \etal\cite{GwakCCGS17} learns shapes from multi-view silhouettes by ray-tracing pooling and further constrains the ill-posed problem using GAN. 
Our approach belongs to this category but is fundamentally different from the existing methods. Rather than sequentially feeding in images, our method learns a GCN to deform the mesh using features pooled from all input images at once. 


\section{Method}

Our model receives multiple color images of an object captured from different viewpoints (with known camera poses) and produces a 3D mesh model in the world coordinate. 
The whole framework adopts the strategy of coarse-to-fine (Fig. \ref{fig:pipeline}), in which a plausible but rough shape is generated first, and details are added later.
Realizing that existing 3D shape generators usually produce reasonable shape even from a single image, we simply use Pixel2Mesh \cite{wang2018pixel2mesh} trained either from single or multiple views to produce the coarse shape, which is taken as input to our Multi-View Deformation Network (MDN) for further improvement.
In MDN, each vertex first samples a set of deformation hypotheses from its surrounding area (Fig. \ref{fig:sampling} (a)). 
Each hypothesis then pools cross-view perceptual feature from early layers of a perceptual network, where the feature resolution is high and contains more low-level geometry information (Fig. \ref{fig:sampling} (b)).
These features are further leveraged by the network to reason the best deformation to move the vertex.
It is worth noting that our MDN can be applied iteratively for multiple times to gradually improve shapes.

\subsection{Multi-View Deformation Network}
In this section, we introduce Multi-View Deformation Network, which is the core of our system to enable the network exploiting cross-view information for shape generation. It first generates deformation hypotheses for each vertex and learns to reason an optimum using feature pooled from inputs.
Our model is essentially a GCN, and can be jointly trained with other GCN based models like Pixel2Mesh.
We refer reader to \cite{BronsteinBLSV17,KipfW16} for details about GCN, and Pixel2Mesh \cite{wang2018pixel2mesh} for graph residual block which will be used in our model.

\vspace{-2mm}
\subsubsection{Deformation Hypothesis Sampling}
\begin{figure}[ht]
	\centering
	\includegraphics[width=\columnwidth]{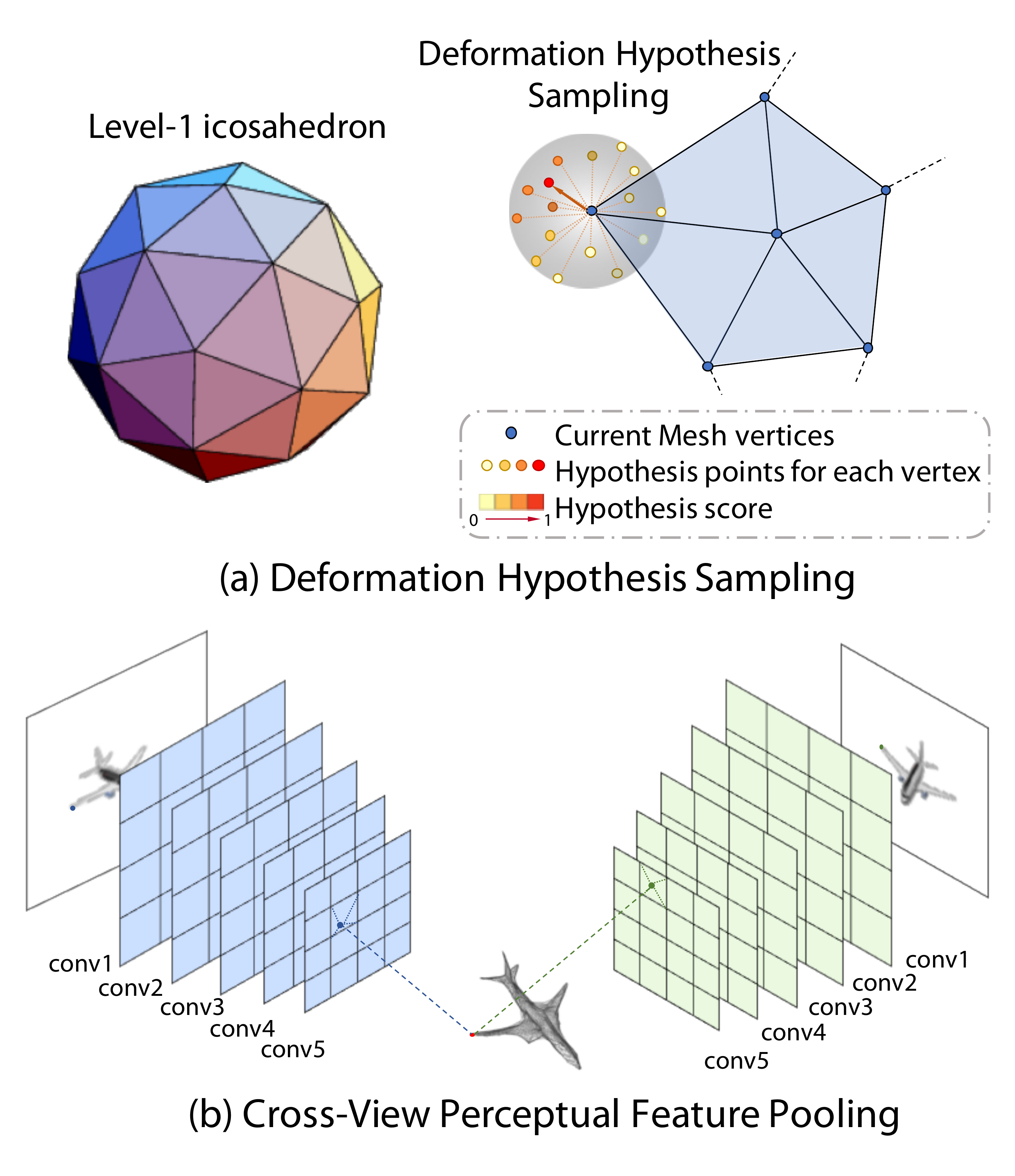}
	\caption{{\bf Deformation Hypothesis and Perceptual Feature Pooling.} (a) Deformation Hypothesis Sampling. We sample 42 deformation hypotheses from a level-1 icosahedron and build a GCN among hypotheses and the vertex. (b) Cross-View Perceptual Feature Pooling. The 3D vertex coordinates are projected to multiple 2D image planes using camera intrinsics and extrinsics. Perceptual features are pooled using bilinear interpolation, and feature statistics are kept on each hypothesis.}
	\label{fig:sampling}
\end{figure}

The first step is to propose deformation hypotheses for each vertex.
This is equivalent as sampling a set of target locations in 3D space where the vertex can be possibly moved to.
To uniformly explore the nearby area, we sample from a level-1 icosahedron centered on the vertex with a scale of 0.02, which results in 42 hypothesis positions (Fig. \ref{fig:sampling} (a), left).
We then build a local graph with edges on the icosahedron surface and additional edges between the hypotheses to the vertex in the center, which forms a graph with 43 nodes and $120+42=162$ edges. 
Such local graph is built for all the vertices, and then fed into a GCN to predict vertex movements (Fig. \ref{fig:sampling} (a), right).

\vspace{-2mm}
\subsubsection{Cross-View Perceptual Feature Pooling}
The second step is to assign each node (in the local GCN) features from the multiple input color images.
Inspired by Pixel2Mesh, we use the prevalent VGG-16 architecture to extract perceptual features.
Since we assume known camera poses, each vertex and hypothesis can find their projections in all input color image planes using known camera intrinsics and extrinsics and pool features from four neighboring feature blocks using bilinear interpolation (Fig. \ref{fig:sampling} (b)).
Different from Pixel2Mesh where high level features from later layers of the VGG (i.e. `conv3\_3', `conv4\_3’, and `conv5\_3’) are pooled to better learn shape priors, MDN pools features from early layers (i.e. `conv1\_2', `conv2\_2’, and `conv3\_3’), which are in high spatial resolution and considered maintaining more detailed information.

To combine multiple features, concatenation has been widely used as a loss-less way, however ends up with total dimension changing with respect to (\wrt) the number of input images.
Statistics feature has been proposed for multi-view shape recognition \cite{su15mvcnn} to handle this problem.
Inspired by this, we concatenate some statistics ($mean$, $max$, and $std$) of the features pooled from all views for each vertex, which makes our network naturally adaptive to variable input views and behave invariant to different input orders.
This also encourages the network to learn from cross-view feature correlations rather than each individual feature vector.
In addition to image features, we also concatenate the 3-dimensional vertex coordinate into the feature vector. 
In total, we compute for each vertex and hypothesis a 1347 dimension feature vector.

\begin{figure*}[t]
	\centering
	\includegraphics[width=\textwidth]{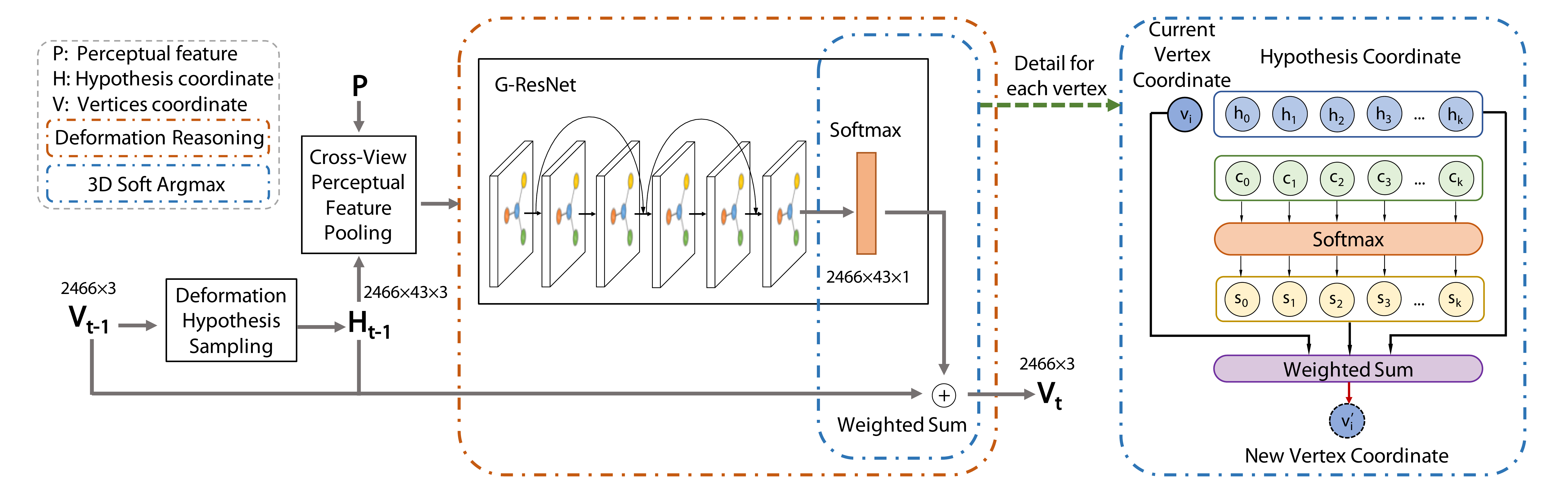}
	\caption{{\bf Deformation Reasoning.} The goal is to reason a good deformation from the hypotheses and pooled features. We first estimate a weight (green circle) for each hypothesis using a GCN. The weights are normalized  by a softmax layer (yellow circle), and the output deformation is the weighted sum of all the deformation hypotheses.}
	\label{fig:ld}
\end{figure*}

\subsubsection{Deformation Reasoning}
The next step is to reason an optimal deformation for each vertex from the hypotheses using pooled cross-view perceptual features.
Note that picking the best hypothesis of all needs an argmax operation, which requires stochastic optimization and usually is not optimal.
Instead, we design a differentiable network component to produce desirable deformation through soft-argmax of the 3D deformation hypotheses, which is illustrated in Fig. \ref{fig:ld}.
Specifically, we first feed the cross-view perceptual feature $P$ into a scoring network, consisting of 6 graph residual convolution layers \cite{wang2018pixel2mesh} plus ReLU, to predict a scalar weight $c_i$ for each hypothesis.
All the weights are then fed into a softmax layer and normalized to scores $s_i$, with $\sum_{i=1}^{43} s_i=1$.
The vertex location is then updated as the weighted sum of all the hypotheses,
i.e. $v=\sum_{i=1}^{43} s_i*h_i$, where $h_i$ is the location of each deformation hypothesis including the vertex itself.
This deformation reasoning unit runs on all local GCN built upon every vertex with shared weights, as we expect all the vertices leveraging multi-view feature in a similar fashion.

\subsection{Loss}
We train our model fully supervised using ground truth 3D CAD models.
Our loss function includes all terms from Pixel2Mesh, but extends the Chamfer distance loss to a re-sampled version.
Chamfer distance measures ``distance'' between two point clouds, which can be problematic when points are not uniformly distributed on the surface.
We propose to randomly re-sample the predicted mesh when calculating Chamfer loss using the re-parameterization trick proposed in Ladický \etal \cite{ladicky2017point}.
Specifically, given a triangle defined by 3 vertices $\left\{ v _ { 1 } , v _ { 2 } , v _ { 3 } \right\} \in \mathbb { R } ^ { 3 }$, a uniform sampling can be achieved by:
$$
s = \left( 1 - \sqrt { r _ { 1 } } \right) v _ { 1 } + \left( 1 - r _ { 2 } \right) \sqrt { r _ { 1 } v _ { 2 } } + \sqrt { r _ { 1 } } r _ { 2 } v _ { 1 },
$$
where $s$ is a point inside the triangle, and $r _ { 1 } , r _ { 2 } \sim U [ 0,1 ]$.
Knowing this, when calculating the loss, we uniformly sample our generated mesh for 4000 points, with the number of points per triangle proportional to its area.
We find this is empirically sufficient to produce a uniform sampling on our output mesh with 2466 vertices, and calculating Chamfer loss on the re-sampled point cloud, containing 6466 in total, helps to remove artifacts in the results.

\subsection{Implementation Details}
For initialization, we use Pixel2Mesh to generate a coarse shape with 2466 vertices.
To improve the quality of initial mesh, we equip the Pixel2Mesh with our cross-view perceptual feature pooling layer, which allows it to extract features from multiple views.

The network is implemented in Tensorflow and optimized using Adam with weight decay as 1e-5 and mini-batch size as 1. 
The model is trained for 50 epochs in total.
For the first 30 epochs, we only train the multi-view Pixel2Mesh for initialization with learning rate 1e-5. Then, we make the whole model trainable, including the VGG for perceptual feature extraction, for another 20 epoch with the learning rate as 1e-6.
The whole model is trained on NVIDIA Titan Xp for 96 hours. 
During training, we randomly pick three images for a mesh as input.
During testing, it takes 0.32s to generate a mesh.

\section{Experiments}
In this section, we perform extensive evaluation of our model for multi-view shape generation.
We compare to state-of-the-art methods, as well as conduct controlled experiments \wrt various aspects, \eg cross category generalization, quantity of inputs, etc.

\subsection{Experimental setup}
\paragraph{Dataset}
We adopt the dataset provided by Choy \etal \cite{choy20163d} as it is widely used by many existing 3D shape generation works. The dataset is created using a subset of ShapeNet\cite{chang2015shapenet} containing 50k 3D CAD models from 13 categories. Each model is rendered from 24 randomly chosen camera viewpoints, and the camera intrinsic and extrinsic parameters are given. For fair comparison, we use the same training/testing split as in Choy \etal \cite{choy20163d} for all our experiments.
\paragraph{Evaluation Metric}
We use standard evaluation metrics for 3D shape generation. Following Fan \etal \cite{fan2017point}, we calculate Chamfer Distance(CD) between points clouds uniformly sampled from the ground truth and our prediction to measure the surface accuracy. We also use F-score following Wang \etal \cite{wang2018pixel2mesh} to measure the completeness and precision of generated shapes. For CD, the smaller is better. For F-score, the larger is better.


\begin{table*}[h]
  \centering
  \begin{tabu} to \textwidth {X[1.5,l]X[1.5,c]X[c]X[1.5,c]X[1.5,c]X[c]X[1.5,c]X[c]X[1.5,c]X[1.5,c]X[c]}
    \toprule
    \multirow{3}{*}{Category} &
    \multicolumn{5}{c}{F-score($\tau$) $\uparrow$}&
    \multicolumn{5}{c}{F-score($2\tau$) $\uparrow$}\\
    \tabuphantomline
    \cmidrule(lr){2-6} \cmidrule(lr){7-11}
    & 3DR2N2$^{\dagger}$ & LSM & MVP2M & P2M-M & Ours & 3DR2N2$^{\dagger}$ & LSM& MVP2M & P2M-M & Ours \\
    \midrule
    \midrule
    Couch & 45.47 & 43.02 & 53.17 & 53.70 &\textbf{57.56}
          & 59.97 & 55.49 & 73.24 & 72.04 & \textbf{75.33}\\
    Cabinet & 54.08 & 50.80 & 56.85 & 63.55 & \textbf{65.72}
            & 64.42 & 60.72 & 76.58 & 79.93 & \textbf{81.57}\\
    Bench & 44.56 & 49.33 & 60.37 & 61.14 & \textbf{66.24}
          & 62.47 & 65.92 & 75.69 & 75.66 & \textbf{79.67}\\
    Chair & 37.62 & 48.55 & 54.19 & 55.89 & \textbf{62.05}
          & 54.26 & 64.95 & 72.36 & 72.36 & \textbf{77.68}\\
    Monitor & 36.33 & 43.65 & 53.41 & 54.50 & \textbf{60.00}
            & 48.65 & 56.33 & 70.63 & 70.51 & \textbf{75.42}\\
    Firearm & 55.72 & 56.14 & 79.67 & 74.85 & \textbf{80.74}
          & 76.79 & 73.89 & 89.08 & 84.82 & \textbf{89.29}\\
    Speaker & 41.48 & 45.21 & 48.90 & 51.61 & \textbf{54.88}
            & 52.29 & 56.65 & 68.29 & 68.53 & \textbf{71.46}\\
    Lamp  & 32.25 & 45.58 & 50.82 & 51.00 & \textbf{62.56}
          & 49.38 & 64.76 & 65.72  & 64.72 & \textbf{74.00}\\
    Cellphone & 58.09 & 60.11 & 66.07 & 70.88 & \textbf{74.36}
              & 69.66 & 71.39 & 82.31 & 84.09 & \textbf{86.16}\\
    Plane & 47.81 & 55.60 & 75.16 & 72.36 & \textbf{76.79}
          & 70.49 & 76.39 & 86.38 & 82.74 & \textbf{86.62}\\
    Table & 48.78 & 48.61 & 65.95 & 67.89 & \textbf{71.89}
          & 62.67 & 62.22 & 79.96 & 81.04 & \textbf{84.19}\\
    Car   & 59.86 & 51.91 & 67.27 & 67.29 & \textbf{68.45}
          & 78.31 & 68.20 & 84.64 & 84.39 & \textbf{85.19}\\
    Watercraft & 40.72 & 47.96 & 61.85 & 57.72 & \textbf{62.99}
               & 63.59 & 66.95 & \textbf{77.49} & 72.96 & 77.32 \\
    \midrule
    Mean & 46.37 & 49.73 & 61.05 & 61.72 & \textbf{66.48}
         & 62.53 & 64.91 & 77.10 & 76.45 & \textbf{80.30}\\
    \bottomrule
  \end{tabu}
  \vspace{1mm}
  \caption{{\bf Comparison to Multi-view Shape Generation Methods.} We show F-score on each semantic category. Our model significantly outperforms previous methods, i.e. 3DR2N2 \cite{choy20163d} and LSM \cite{kar2017learning}, and competitive baselines derived from Pixel2Mesh \cite{wang2018pixel2mesh}. Please see supplementary materials for Chamfer Distance. The notation $\dagger$ indicates the methods which does not require camera extrinsics.}
  \label{tb1:multiview}
\end{table*}

\subsection{Comparison to Multi-view Shape Generation}
\label{sec:mvp2m}
We compare to previous works for multi-view shape generation and show effectiveness of MDN in improving shape quality.
While most shape generation methods take only a single image, we find Choy \etal \cite{choy20163d} and Kar \etal \cite{kar2017learning} work in the same setting with us.
We also build two competitive baselines using Pixel2Mesh.
In the first baseline (Tab.\ref{tb1:multiview}, P2M-M), we directly run single-view Pixel2Mesh on each of the input image and fuse multiple results \cite{curless1996volumetric,LorensenC87}.
In the second baseline (Tab.\ref{tb1:multiview}, MVP2M), we replace the perceptual feature pooling to our cross-view version to enable Pixel2Mesh for the multi-view scenario (more details in supplementary materials).

Tab. \ref{tb1:multiview} shows quantitative comparison in F-score. 
As can be seen, our baselines already outperform other methods, which shows the advantage of mesh representation in finding surface details.
Moreover, directly equipping Pixel2Mesh with multi-view features does not improve too much (even slightly worse than the average of multiple runs of single-view Pixel2Mesh), which shows dedicate architecture is required to efficiently learn from multi-view features.
In contrast, our Multi-View Deformation Network significantly further improves the results from the MVP2M baseline (i.e. our coarse shape initialization).

More qualitative results are shown in Fig. \ref{fig:multiview}.
We show results from different methods aligned with one input view (left) and a random view (right).
As can be seen, Choy \etal \cite{choy20163d} (3D-R2N2) and Kar \etal \cite{kar2017learning} (LSM) produce 3D volume, which lose thin structures and surface details.
Pixel2Mesh (P2M) produces mesh models but shows obvious artifacts when visualized in viewpoint other than the input.
In comparison, our results contain better surface details and more accurate geometry learned from multiple views.

\subsection{Generalization Capability}
Our MDN is inspired by multi-view geometry methods, where 3D location is reasoned via cross-view information.
In this section, we investigate the generalization capability of MDN in many aspects to improve the initialization mesh. For all the experiments in this section, we fix the coarse stage and train/test MDN under different settings.

\begin{figure}
\scalebox{0.85}{
    \centering{}%
    \begin{tabular}{cc}
    \hspace{-4mm}
    \begin{tabular}{c}
    {\small{}}%
    \begin{tabular}{ccc}
    \toprule
    {\small{}Category } & {\small{}Except } & {\small{}All}\tabularnewline
    \hline 
    \hline 
    {\small{}lamp } & {\small{}10.96} & {\small{}11.73}\\
    {\small{}cabinet } & {\small{}8.88} & {\small{}8.99}\\
    {\small{}cellphone } & {\small{}7.10} & {\small{}8.29}\\
    {\small{}chair } & {\small{}6.49} & {\small{}7.86}\\
    {\small{}monitor } & {\small{}6.06} & {\small{}6.60}\\
    {\small{}speaker } & {\small{}5.75} & {\small{}5.98}\\
    {\small{}table } & {\small{}5.44} & {\small{}5.94}\\
    {\small{}bench } & {\small{}5.30} & {\small{}5.87}\\
    {\small{}couch } & {\small{}3.76} & {\small{}4.39}\\
    {\small{}plane } & {\small{}1.12} & {\small{}1.63}\\
    {\small{}firarm } & {\small{}0.67} & {\small{}1.07}\\
    {\small{}watercraft } & {\small{}0.21} & {\small{}1.14}\\
    {\small{}car } & {\small{}0.14} & {\small{}1.18}\\
    \bottomrule
    \end{tabular}\tabularnewline
    \end{tabular} &
    \hspace{-0.35in}%
    \begin{tabular}{c}
    \includegraphics[width=0.3\textwidth]{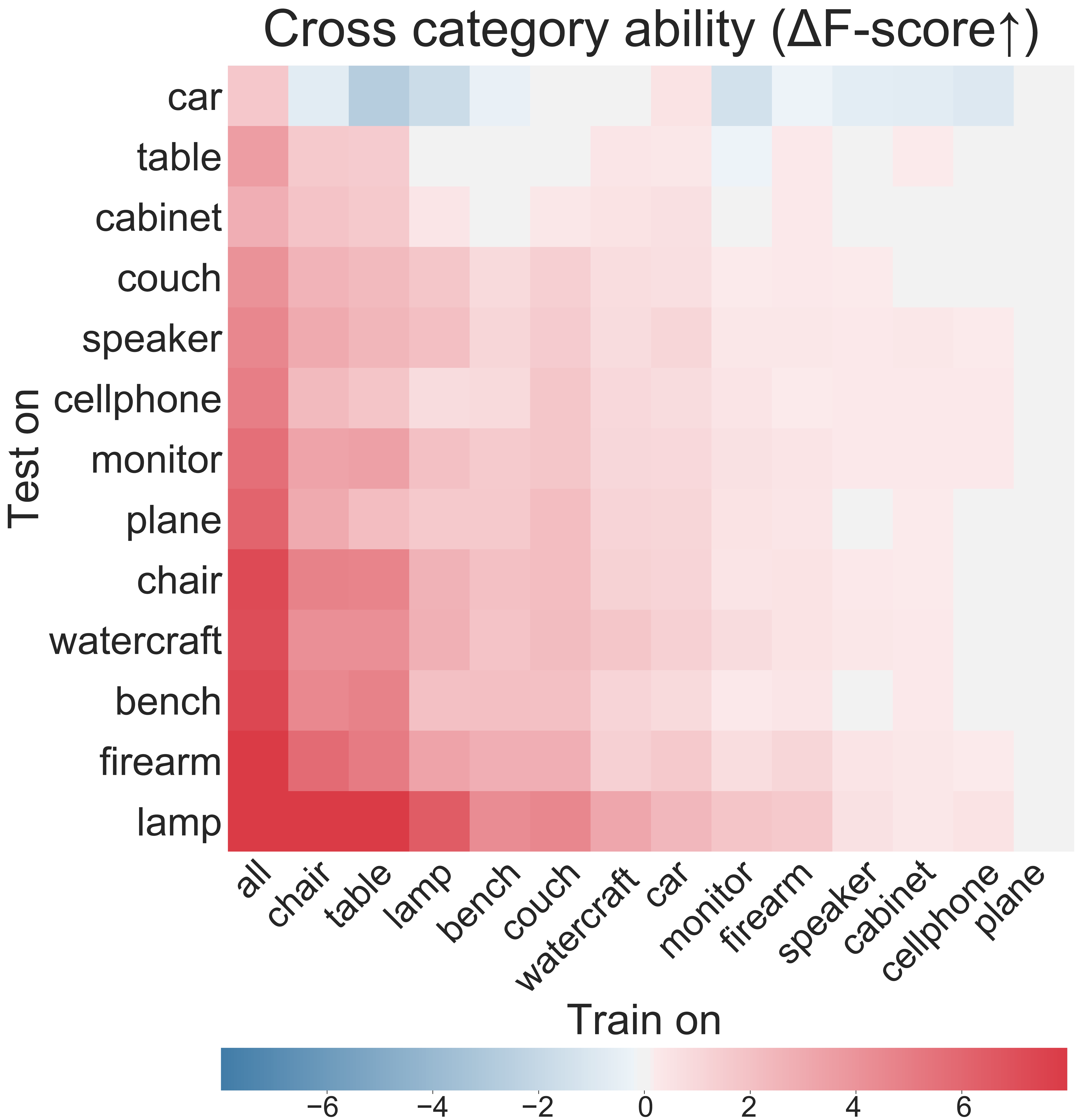}\tabularnewline
    \end{tabular} 
    \tabularnewline
    {\small{}(a) Train except one category} & {\small{}(b)Train on one category}\tabularnewline
    \end{tabular}
}
    \vspace{1mm}
    \caption{{\bf Cross-Category Generalization.} (a) MDN trained on 12 out of 13 categories and tested on the one left. (b) MDN trained on 1 category and tested on the other. Each block represents the experiment with MDN trained on horizontal category and tested on vertical category. Both (a) and (b) show improvements of F-score($\tau$) upon MVP2M through MDN.}
    \label{fig:cross}
\end{figure}

\begin{figure*}[t]
	\centering
	\includegraphics[width=\textwidth]{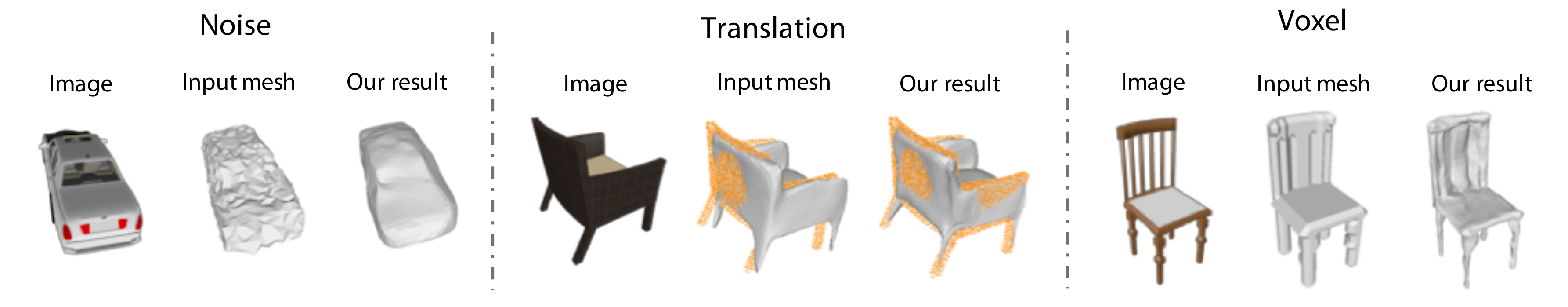}
	\caption{{\bf Robustness to Initialization.} Our model is robust to added noise, shift, and input mesh from other sources.}
	\label{fig:init}
\end{figure*}
\nopagebreak
\subsubsection{Semantic Category}
We first verify how our network generalizes across semantic categories.
We fix the initial MVP2M and train MDN with 12 out 13 categories and test on the one left out, and the improvements upon initialization are shown in Fig. \ref{fig:cross} (a). 
As can be seen, the performance is only slightly lower when the testing category is removed from the training set compared to the model trained using all categories.
To make it more challenging, we also train MDN on only one category and test on all the others.
Surprisingly, MDN still generalizes well between most of the categories as shown in Fig. \ref{fig:cross} (b).
Strong generalizing categories (e.g. chair, table, lamp) tend to have relatively complex geometry, thus the model has better chance to learn from cross-view information.
On the other hand, categories with super simple geometry (e.g. speaker, cellphone) do not help to improve other categories, even not for themselves.
On the whole, MDN shows good generalization capability across semantic categories.

\subsubsection{Number of Views}
We then test how MDN performs \wrt the number of input views.
In Tab. \ref{tbl:view_number}, we see that MDN consistently performs better when more input views are available, even though the number of view is fixed as 3 for efficiency during the training.
This indicates that features from multiple views are well encoded in the statistics, and MDN is able to exploit additional information when seeing more images.
For reference, we train five MDNs with the input view number fixed at 2 to 5 respectively. 
As shown in Tab. \ref{tbl:view_number} ``Resp.'', the 3-view MDN performs very close to models trained with more views (e.g. 4 and 5), which shows the model learns efficiently from fewer number of views during the training.
The 3-view MDN also outperform models trained with less views (e.g. 2), which indicates additional information provided during the training can be effectively activated during the test even when observation is limited.
Overall, MDN generalizes well to different number of inputs.

\begin{table}[h]
    \begin{center}
    \resizebox{\columnwidth}{!}{
        \begin{tabular}{cccccc}
        \toprule
        \#train & \#test & 2 & 3 & 4 & 5 \\
        \midrule
        \multirow{3}{*}{3} & F-score($\tau$) $\uparrow$ & 64.48 & 66.44 & 67.66 & 68.29 \\
        & F-score($2 \tau$) $\uparrow$ & 78.74 & 80.33 & 81.36 & 81.97 \\
        & CD $\downarrow$ & 0.515 & 0.484 & 0.468 & 0.459\\
        \midrule
        \multirow{3}{*}{Resp.} & F-score($\tau$) $\uparrow$& 64.11 & 66.44 & 68.54 & 68.82 \\
        & F-score($2 \tau$) $\uparrow$ & 78.34 & 80.33 & 81.56 & 81.99 \\
        & CD $\downarrow$ & 0.527 & 0.484 & 0.467 & 0.452\\
        \bottomrule
        \end{tabular}
        }
    \end{center}
    \caption{{\bf Performance \wrt Input View Numbers.} Our MDN performs consistently better when more view is given, even trained using only 3 views.}
\label{tbl:view_number}
\vspace{-8mm}
\end{table}

\subsubsection{Initialization}
Lastly, we test if the model overfits to the input initialization, i.e. the MVP2M.
To this end, we add translation and random noise to the rough shape from MVP2M. 
We also take as input the mesh converted from 3DR2N2 using marching cube \cite{LorensenC87}.
As shown in Fig. \ref{fig:init}, MDN successfully removes the noise, aligns the input with ground truth, and adds significant geometry details.
This shows that MDN is tolerant to input variance.

\subsection{Ablation Study}
In this section, we verify the qualitative and quantitative improvements from statistic feature pooling, re-sampled Chamfer distance, and iterative refinement. 

\begin{figure}[t]
	\centering
	\includegraphics[width=\columnwidth]{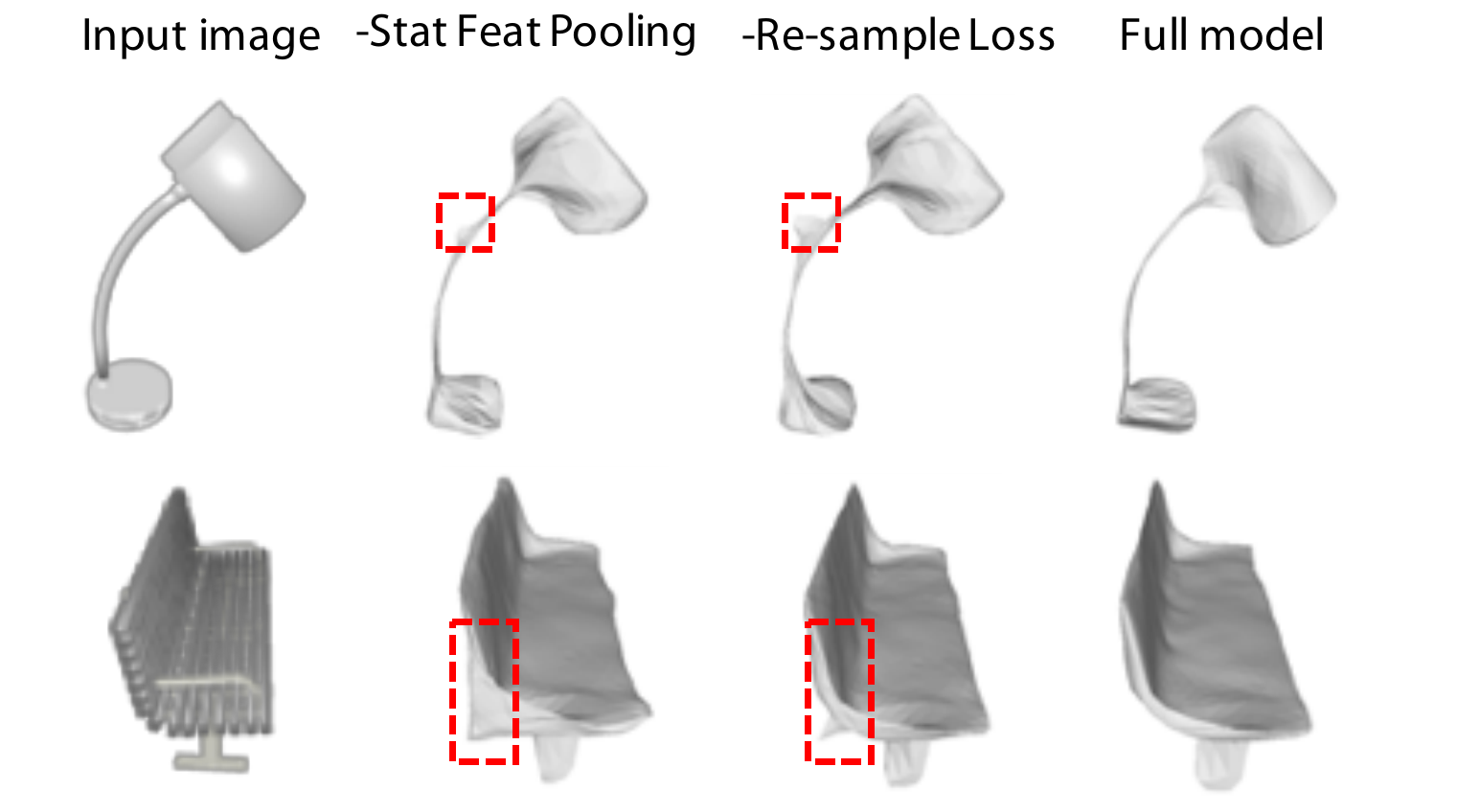}
	\caption{{\bf Qualitative Ablation Study.} We show meshes from the MDN with statistics feature or re-sampling loss disabled.}
	\label{Fig.ablation}
\end{figure}

\begin{table}[t]
    \centering
    \begin{tabular}{cccc}
        \toprule
        Metrics & F-score($\tau$) $\uparrow$ & F-score($2\tau$) $\uparrow$ & CD $\downarrow$\\
        \midrule
        -Feat Stat & 65.26 & 79.13 & 0.511\\
        -Re-sample Loss & 66.26 & 80.04 & 0.496\\
        Full Model & \textbf{66.48} & \textbf{80.30} & \textbf{0.486} \\
        \bottomrule
    \end{tabular}
    \vspace{5pt}
    \caption{{\bf Quantitative Ablation Study.} We show the metrics of the MDN with statistics feature or re-sampling loss disabled.}
    \label{tbl:ablation}
\end{table}

\subsubsection{Statistical Feature}
We first check the importance of using feature statistics. We train MDN with the ordinary concatenation. This maintains all the features loss-less to potentially produce better geometry, but does not support variable number of inputs any more.
Surprisingly, our model with feature statistics (Tab. \ref{tbl:ablation}, ``Full Model'') still outperforms the one with concatenation (Tab. \ref{tbl:ablation}, ``-Feat Stat'').
This is probably because that our feature statistics is invariant to the input order, such that the network learns more efficiently during the training. It also explicitly encodes cross-view feature correlations, which can be directly leveraged by the network.

\begin{figure*}[th]
	\centering
	\vspace{-2mm}
	\includegraphics[width=\textwidth]{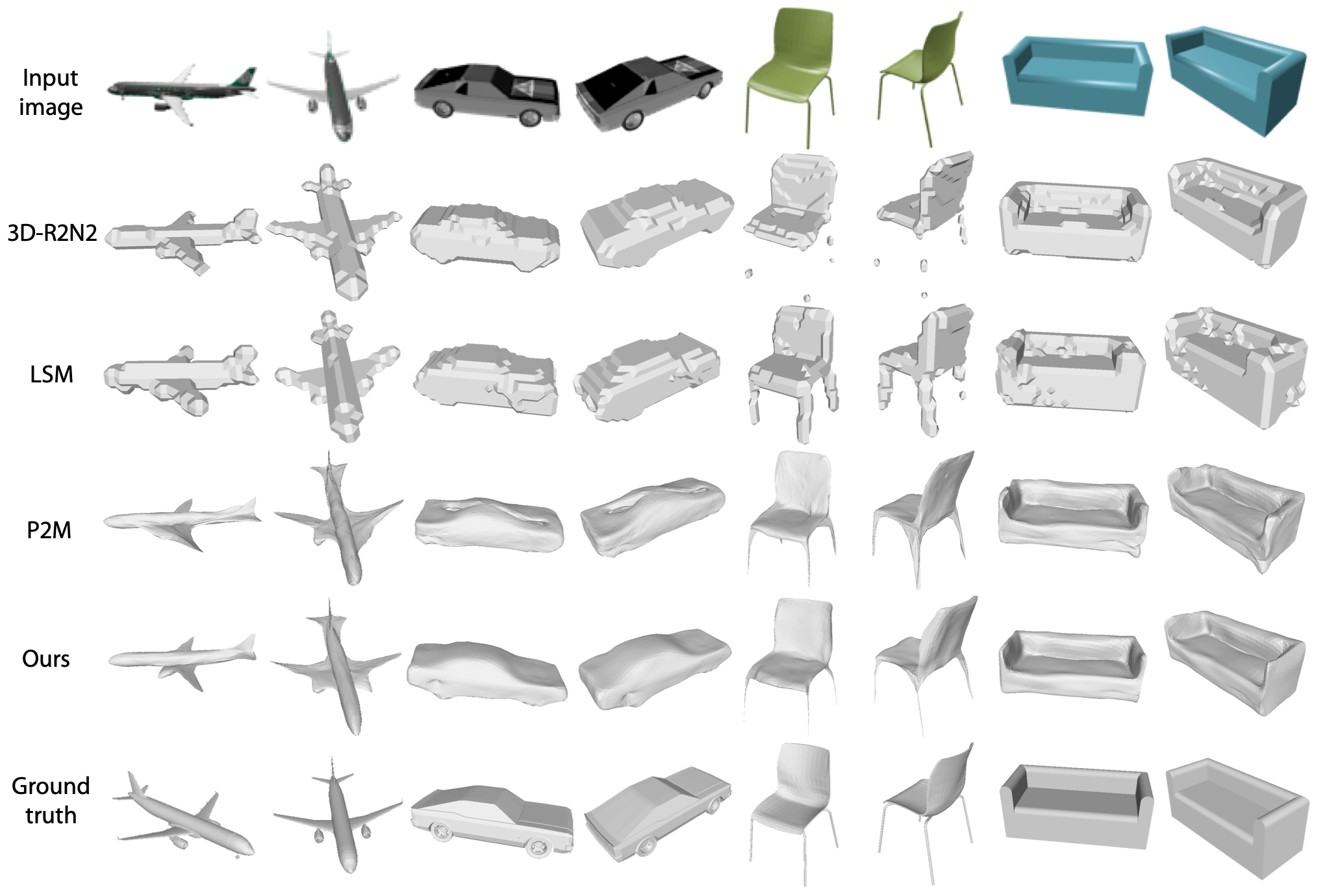}
	\caption{{\bf Qualitative Evaluation.} From top to bottom, we show in each row: two camera views, results of 3DR2N2, LSM, multi-view Pixel2Mesh, ours, and the ground truth. Our predicts maintain good details and align well with different camera views.
	Please see supplementary materials for more results. 
	}
	\label{fig:multiview}
	\vspace{-2mm}
\end{figure*}

\subsubsection{Re-sampled Chamfer Distance}
We then investigate the impact of the re-sampled Chamfer loss.
We train our model using the traditional Chamfer loss only on mesh vertices as defined in Pixel2Mesh, and all metrics drop consistently (Tab. \ref{tbl:ablation}, ``-Re-sample Loss''). 
Intuitively, our re-sampling loss is especially helpful for places with sparse vertices and irregular faces, such as the elongated lamp neck as shown in Fig. \ref{Fig.ablation}, 3rd column.
It also prevents big mistakes from happening on a single vertex, e.g. the spike on bench, where our loss penalizes a lot of sampled points on wrong faces caused by the vertex but the standard Chamfer loss only penalizes one point.


\begin{figure}[t]
	\centering
	\includegraphics[width=\columnwidth]{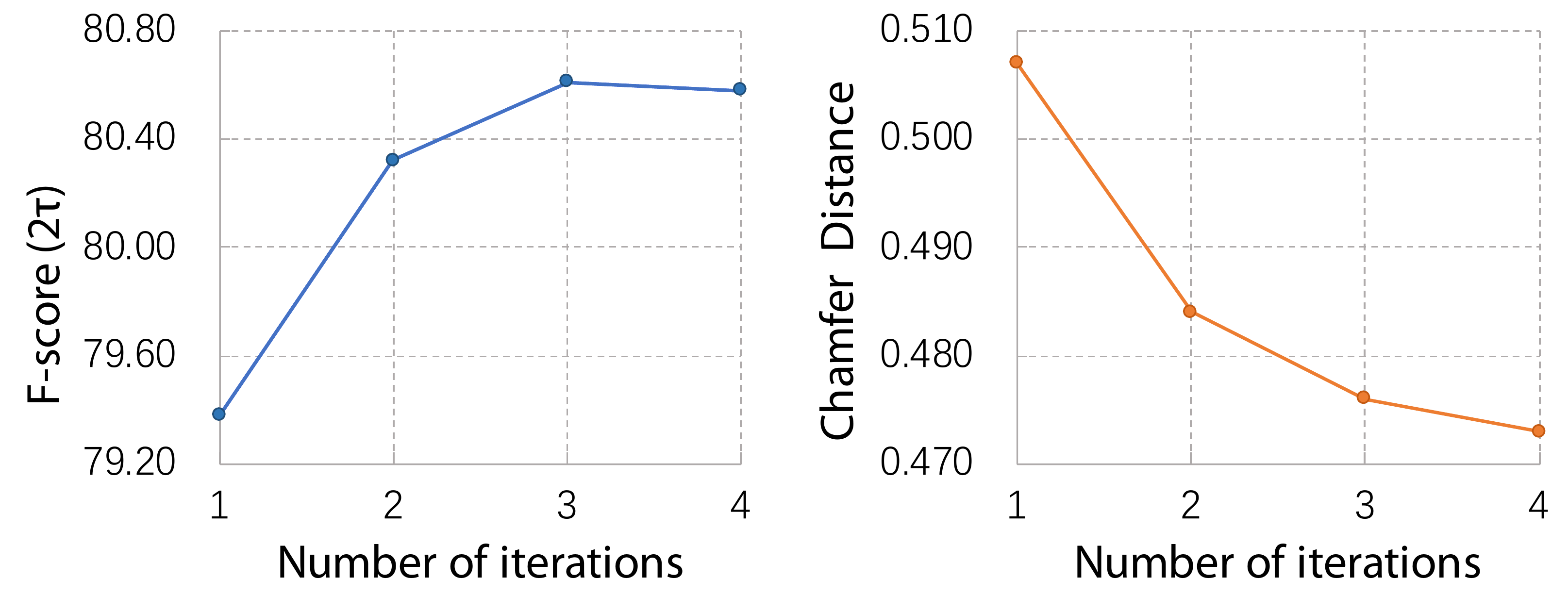}
	\caption{{\bf Performance with Different Iterations. } The performance keeps improving with more iterations and saturate at three.
    }
	\label{fig:iter}
	\vspace{-4mm}
\end{figure}

\subsubsection{Number of Iteration}
\vspace{-1mm}
Figure \ref{fig:iter} shows that the performance of our model keeps improving with more iterations, and is roughly saturated at three.
Therefore we choose to run three iterations during the inference even though marginal improvements can be further obtained from more iterations.

\section{Conclusion}
We propose a graph convolutional framework to produce 3D mesh model from multiple images.
Our model learns to exploit cross-view information and generates vertex deformation iteratively to improve the mesh produced from the direct prediction methods, e.g. Pixel2Mesh and its multi-view extension.
Inspired by multi-view geometry methods, our model searches in the nearby area around each vertex for an optimal place to relocate it.
Compared to previous works, our model achieves the state-of-the-art performance, produces shapes containing accurate surface details rather than merely visually plausible from input views, and shows good generalization capability in many aspects.
For future work, combining with efficient shape retrieval for initialization, integrating with multi-view stereo models for explicit photometric consistency, and extending to scene scales are some of the practical directions to explore.
On a high level, how to integrating the similar idea in emerging new representations, such as part based model with shape basis and learned function \cite{Park_2019_CVPR} are interesting for further study.


\clearpage

{\small
\bibliographystyle{ieee_fullname}
\bibliography{egbib}
}

\clearpage
\renewcommand\thesection{\Alph{section}}
\setcounter{section}{0}
\begin{widetext}
\begin{center}
\textbf{\Large Supplementary Materials}
\end{center}
\end{widetext}

This supplementary material includes the implementation details, baseline algorithm designs, and more experiment results. 

\section{Network Architecture}
\subsection{Deformation Sampling}
In particular, a level-K icosahedron is obtained by sampling the middle points of all the edges from a level-(K-1) icosahedron, and a level-0 icosahedron vertex coordinates can be calculated as
\begin{equation}
C_0 = \begin{vmatrix}
\phi & 1 & 0\\ 
-\phi & 1 & 0 \\ 
\phi & -1 & 0\\ 
-\phi & -1 & 0\\
1 & 0 & \phi \\
1 & 0 & -\phi \\
-1 & 0 & \phi \\
-1 & 0 & -\phi \\
0 & \phi & 1 \\
0 & -\phi & 1 \\
0 & \phi & -1 \\
0 & -\phi & -1 \\
\end{vmatrix}/\sqrt{1+\phi^{2}}\label{eq:c0},
\end{equation}
where 
\begin{equation}
\phi =\frac{1+\sqrt{5}}{2}.
\end{equation}
Here, each row represents a vertex coordinate on the level-0 icosahedron.

In our experiment, we use Meshlab \cite{cignoni2008meshlab} to generate level-1 icosahedron \cite{wiki:Pentakis_icosidodecahedron} vertices coordinates and corresponding edges for deformation hypotheses and local GCN graph topology.
The vertex coordinates are scaled along the radius to a pre-defined value. More details about connection and implementation can be found in \cite{icosahedron2sphere, xiao2009image}.

\subsection{Deformation Reasoning}
The network architecture for deformation reasoning is shown in Tab. \ref{tab:network}.
The Deformation Reasoning component takes current vertices coordinates, hypothesis feature and hypothesis coordinates as input. It consists of 6 graph convolution layers with residual connections. The last graph convolution layer is followed by a softmax layer and the final output is normalized to weights, which are used to obtain new coordinates for the vertices through weighted sums. 

\begin{table*}

\centering

\begin{tabular}{ccccc}
\toprule 
Tensor Name & Layer & Parameters & Input Tenstor & Activation\tabularnewline
\midrule
\midrule
Vertices Coordinate & Input & - & - & -\tabularnewline
Hypothesis Coordinate & Input & - & - & -\tabularnewline
Hypothesis Feature & Input & - & - & -\tabularnewline
\hline 
GraphConv1 & GraphConv & $339\times 192$ & Hypothesis Feature & ReLU\tabularnewline
GraphConv2 & GraphConv & $192\times 192$ & GraphConv1 & ReLU\tabularnewline
GraphConv3 & GraphConv & $192\times 192$ & GraphConv2 & ReLU\tabularnewline
Add1 & Add & - & GraphConv2, GraphConv3 & -\tabularnewline
GraphConv4 & GraphConv & $192\times 192$ & Add1 & ReLU\tabularnewline
GraphConv5 & GraphConv & $192\times 192$ & GraphConv4 & ReLU\tabularnewline
Add2 & Add & - & GraphConv4, GraphConv5 & -\tabularnewline
GraphConv6 & GraphConv & $192\times 1$ & Add2 & ReLU\tabularnewline
Hypothesis Score & Softmax & - & GraphConv6 & -\tabularnewline
\hline 
New Vertices Coordinate & Weighted Sum & - & \tabincell{c}{Hypothesis Score,\\ Hypothesis Coordinate, Vertices Coordinate} & -\tabularnewline
\bottomrule 
\end{tabular}
\vspace{5pt}
\caption{{\bf Network Architecture for Deformation Reasoning.}}
\label{tab:network}
\end{table*}

\subsection{Perceptual Feature Pooling}
\label{sec:featurepooling}
The perceptual feature pooling layer projects all 3D vertices onto feature maps and obtain the vertices features from the corresponding 2D coordinates. Suppose a 3D vertex with coordinate $(X, Y, Z)$ in the camera view; its 2D projection in image is:
\begin{equation}
    \begin{aligned} x & = \frac { X } { Z } * f _ { x } + c _ { x }, \\ y & = \frac { Y } { Z } * f _ { y } + c _ { y }, \end{aligned}
\end{equation}
where $f_x$ and $f_y$ denote the focal lengths along horizontal and vertical image axis, and $(c_x , c_y)$ is the projection of the camera center.

To pool feature from multiple images for each vertex, we transform the vertex into the camera coordinate of each input views using the camera extrinsic matrix.
Suppose the $\{R,T\}$ is the transformation from the world coordinate to a camera coordinate, and $V$ is the coordinate of a vertex in the world coordinate, its location in the camera coordinate can be obtained by $V_c = R \cdot V+T$.


\section{Baselines Methods}
In Sec. 4.2 of the main submission, we propose two baselines extending the Pixel2Mesh architecture for multi-view shape generation. Here we show more details about them.
\subsection{P2M-M}
For the P2M-M baseline, we first run the single-view Pixel2Mesh on each of the input views to generate a shape respectively.
These shapes are then transformed into the world coordinate and converted into signed distance function (SDF) \cite{curless1996volumetric,Stutz2018CVPR,Stutz2017}.
We directly average these SDFs and run Lorensen \etal \cite{LorensenC87} to obtain the triangular meshes.

\begin{figure}[h]
	\centering
	\includegraphics[width=\columnwidth]{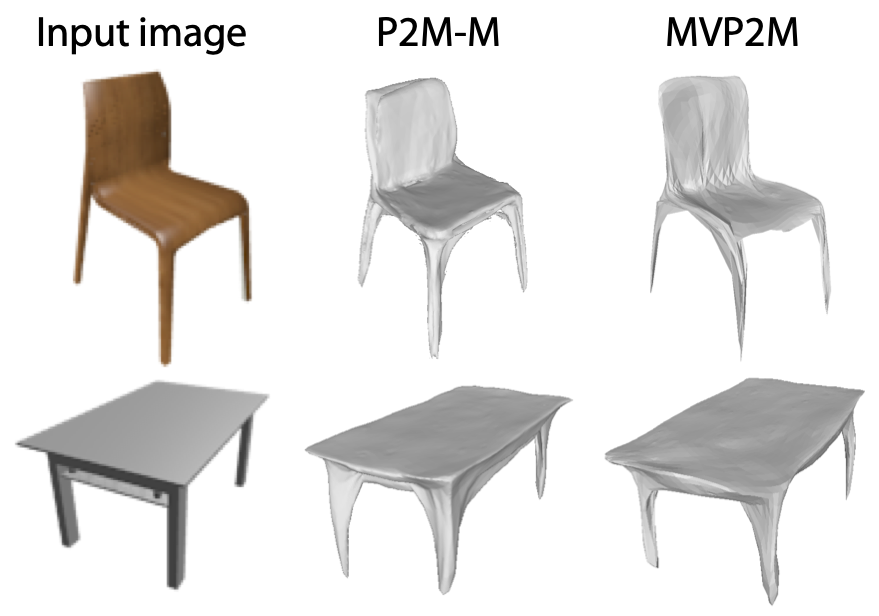}
	\caption{{\bf Results of baselines.}}
	\label{fig:p2m-m}
\end{figure}

\subsection{MVP2M}
For the P2M-M baseline, Pixel2Mesh sees only one image at once.
Here, we extend Pixel2Mesh to access multiple images in a single network forward pass by having it pools multi-view features from all the inputs.
This can be achieved by replacing the perceptual feature pooling layers with our multi-view version as introduced in Sec. \ref{sec:featurepooling}.
In particular, perceptual features are pooled from layer `conv3\_3', `conv4\_3', and `conv5\_3' from the VGG-16 network, and feature statistics (Sec. 3.1.2 in the main submission) are calculated and concatenated, which ends up with a 1280 dimension feature vector. 
In practice, we also tried to pool geometry related features from early convolution layers (i.e. `conv1\_2', `conv2\_2', and `conv3\_3'), but found it doesn't work as well as the case with semantic feature pooled from later layers.

Fig. \ref{fig:p2m-m} shows some examples of results from both baselines.

\section{More Experiment Results}
In this section, we provide more results for quantitative and qualitative evaluations and ablation study.

\subsection{Comparison to State-of-the-art}
In the main submission, we compare to the state-of-the-art methods in F-score. Here we show the comparison in Chamfer distance in Tab. \ref{tb1:multiview}.
Again, we achieve overall the best performance (i.e. the lowest Chamfer distance) comparing to all the previous methods and baselines.
We also achieve the best performance for most of the categories, except for very few categories in which geometry and texture are usually too simple to learn cross-view information.

\begin{table*}[h]
  \centering
  \begin{tabu} to \textwidth {X[1.5,l]X[1.5,c]X[c]X[1.5,c]X[1.5,c]X[1,c]}
    \toprule
    \multirow{2}{*}{Category} &
    \multicolumn{5}{c}{Chamfer Distance(CD) $\downarrow$}\\
    \tabuphantomline
    \cmidrule(lr){2-6}
    & 3DR2N2$^{\dagger}$ & LSM & MVP2M & P2M-M & Ours\\
    \midrule
    \midrule
    Couch & 0.806 & 0.730 & 0.534 & 0.496 &\textbf{0.439}\\
    Cabinet & 0.613 & 0.634 & 0.488 & 0.359 & \textbf{0.337}\\
    Bench & 1.362 & 0.572 & 0.591 & 0.594 & \textbf{0.549}\\
    Chair & 1.534 & 0.495 & 0.583 & 0.561 & \textbf{0.461}\\
    Monitor & 1.465 & 0.592 & 0.658 & 0.654 & \textbf{0.566}\\
    Firearm & 0.432 & 0.385 & 0.305 & 0.428 & \textbf{0.305}\\
    Speaker & 1.443 & 0.767 & 0.745 & 0.697 & \textbf{0.635}\\
    Lamp  & 6.780 & 1.768 & \textbf{0.980} & 1.184 & 1.135\\
    Cellphone & 1.161 & 0.362 & 0.445 & 0.360 & \textbf{0.325}\\
    Plane & 0.854 & 0.496 & \textbf{0.403} & 0.457 & 0.422\\
    Table & 1.243 & 0.994 & 0.511 & 0.441 & \textbf{0.388}\\
    Car   & 0.358 & 0.326 & 0.321 & 0.264 & \textbf{0.249}\\
    Watercraft & 0.869 & 0.509 & \textbf{0.463} & 0.627 & 0.508 \\
    \midrule
    Mean & 1.455 & 0.664 & 0.541 & 0.548 & \textbf{0.486}\\
    \bottomrule
  \end{tabu}
  \vspace{1mm}
  \caption{{\bf Comparison to Multi-view Shape Generation Methods.} We show the Chamfer Distance on each semantic category. Our method achieves the best performance overall. The notation $\dagger$ indicates the methods which does not require camera extrinsics.}
  \label{tb1:multiview}
\end{table*}

\subsection{Ablation Study}

\subsubsection{Effect of Re-sample Loss}
\begin{figure*}[h]
	\centering
	\includegraphics[width=\textwidth]{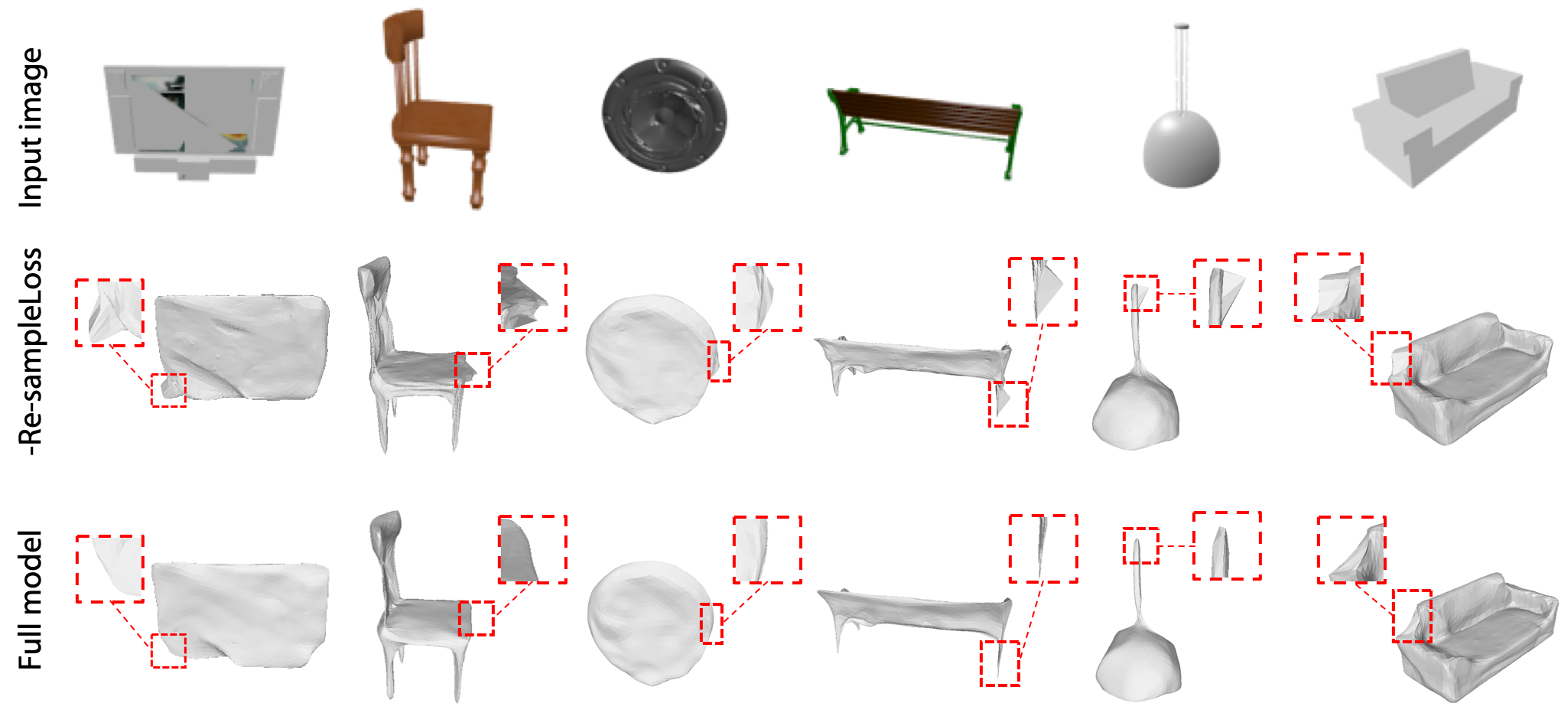}
	\caption{\textbf{Effect of Re-sampled Loss.} We show more qualitative comparison between model trained with original Chamfer loss and our re-sampled version. The re-sampled loss (Full model) helps to prevent the flying pixel and spike artifacts.}
	\label{fig:resample}
\end{figure*}

More comparison between the model trained with the traditional and our re-sampled Chamfer loss is shown in Fig. \ref{fig:resample}. 
As can be seen in the zoom-in areas, our re-sampled Chamfer loss can effectively penalize large flying triangles caused by a few flying vertices
, and thus the results of our full model are free from such artifacts.

\subsubsection{Effect of More Iterations}
In our main submission, we show the numerical improvements with more iterations. Here we show some qualitative results in Fig. \ref{fig:iter}.
As can be seen, thin structure and surface details are recovered throughout iterations as reflected in the zoom-in regions.
\begin{figure*}[h]
	\centering
	\includegraphics[width=\textwidth]{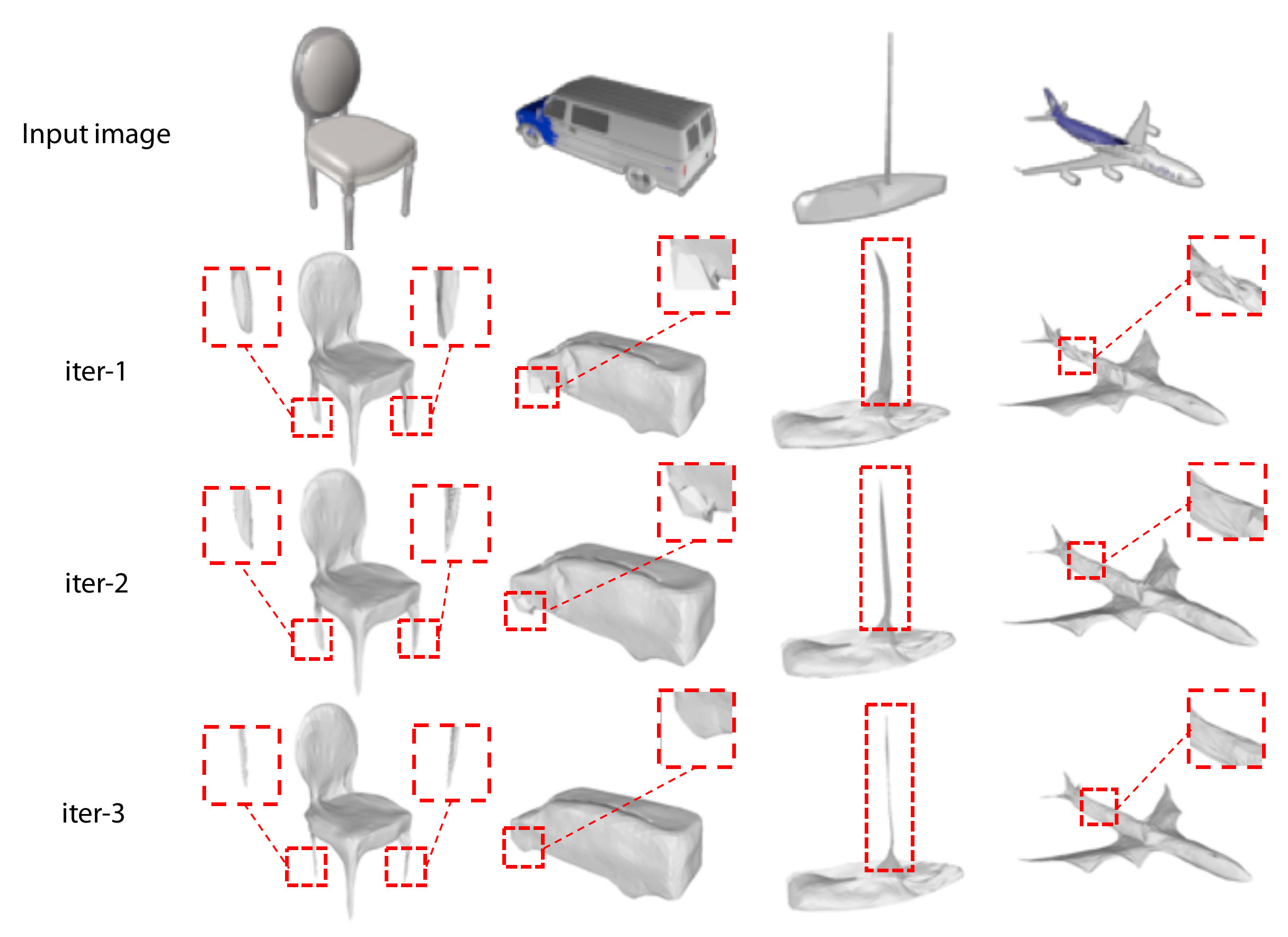}
	\caption{\textbf{Effect of Iterations.} We show the output of our system after each iterations. Thin structures and geometry details are recovered in the later iterations.}
	\label{fig:iter}
\end{figure*}

\subsubsection{Effect of Different Coarse Shape Generation}
We add another experiment using MVP2M and P2M-M as coarse shape initialization methods respectively. Here we show the comparison results in Tab. \ref{tbl:coarse}. MDN consistently improves upon both P2M-M and MVP2M. Ours is also slightly better than P2M-M+MDN as the initialization is better. 
In order to emphasize the generalization ability of using non-ellipsoid initial, we also add experimental results of training on the chair class using other voxel-based methods e.g. 3DR2N2 as a rough shape initialization methods. The qualitative and quantitative result are shown in Fig. \ref{fig:cross}.
As can be seen, MDN generalizes to meshes obtained from 3DR2N2 directly without finetuning.

\begin{table}[h]
    \begin{center}
    \begin{tabu}to \columnwidth{X[2.4c]X[0.8c]X[2c]X[2.1c]}
    \toprule
    {Methods} & {CD$\downarrow$} & {F-score($\tau$)$\uparrow$}& {F-score($2\tau$)$\uparrow$}\tabularnewline
    \hline \hline
    {P2M-M} & {0.548} & {61.72} & {76.45}\\
    {P2M-M+MDN} & {0.493}& {64.47} & {79.31}\\
    \hline
    {MVP2M} & {0.541}& {61.05} & {77.10}\\
    {Ours} & {0.486} & {66.48} & {80.30}\\
    \bottomrule
    \end{tabu}
    \end{center}
    \caption{\textbf{Effect of Different Coarse Shape Generation.}}
\label{tbl:coarse}
\end{table}

\begin{figure}
\scalebox{0.85}{
    \centering{}%
    \begin{tabular}{cc}
    \hspace{-6mm}
    \begin{tabular}{p{3.5cm}}
    {\small{}}
    \centering\includegraphics[width=0.12\textwidth]{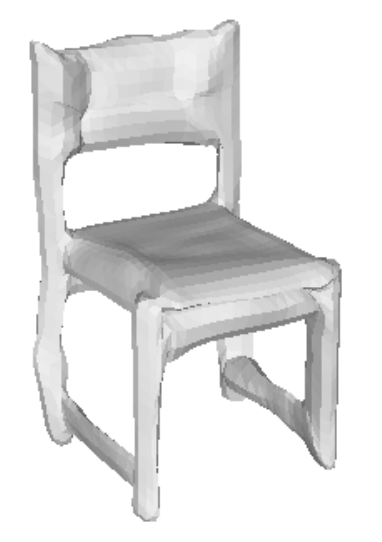}\tabularnewline
    {\small{}(a) Example mesh result.}\tabularnewline
    \end{tabular}&
    \hspace{-0.2in}
    \begin{tabular}{c}
            \begin{tabu}to 0.7\columnwidth{X[c]X[c]X[c]}
                \toprule
                {Metrics} & {w/o MDN} & {w/ MDN}\tabularnewline
                \hline\hline
                {CD} & 2.438 & \textbf{1.418}\\
                {\small{}F-score($\tau$)} & 20.24 & \textbf{36.81}\\
                {\small{}F-score($2\tau$)} & 31.62 & \textbf{52.45}\\
                \bottomrule
            \end{tabu}\tabularnewline
            {\small{}(b) 3DR2N2 scheme with/wo MDN on chair}\tabularnewline
    \end{tabular}
    \end{tabular}
}
    \vspace{0.5mm}
    \caption{ Experiments results using non-ellipsoid initial.}
    \label{fig:cross}
    \vspace{-2em}
\end{figure}

\subsection{More Qualitative Results}
\begin{figure*}[h!]
	\centering
	\includegraphics[width=\textwidth, 
                     height=\textheight, 
                     keepaspectratio]{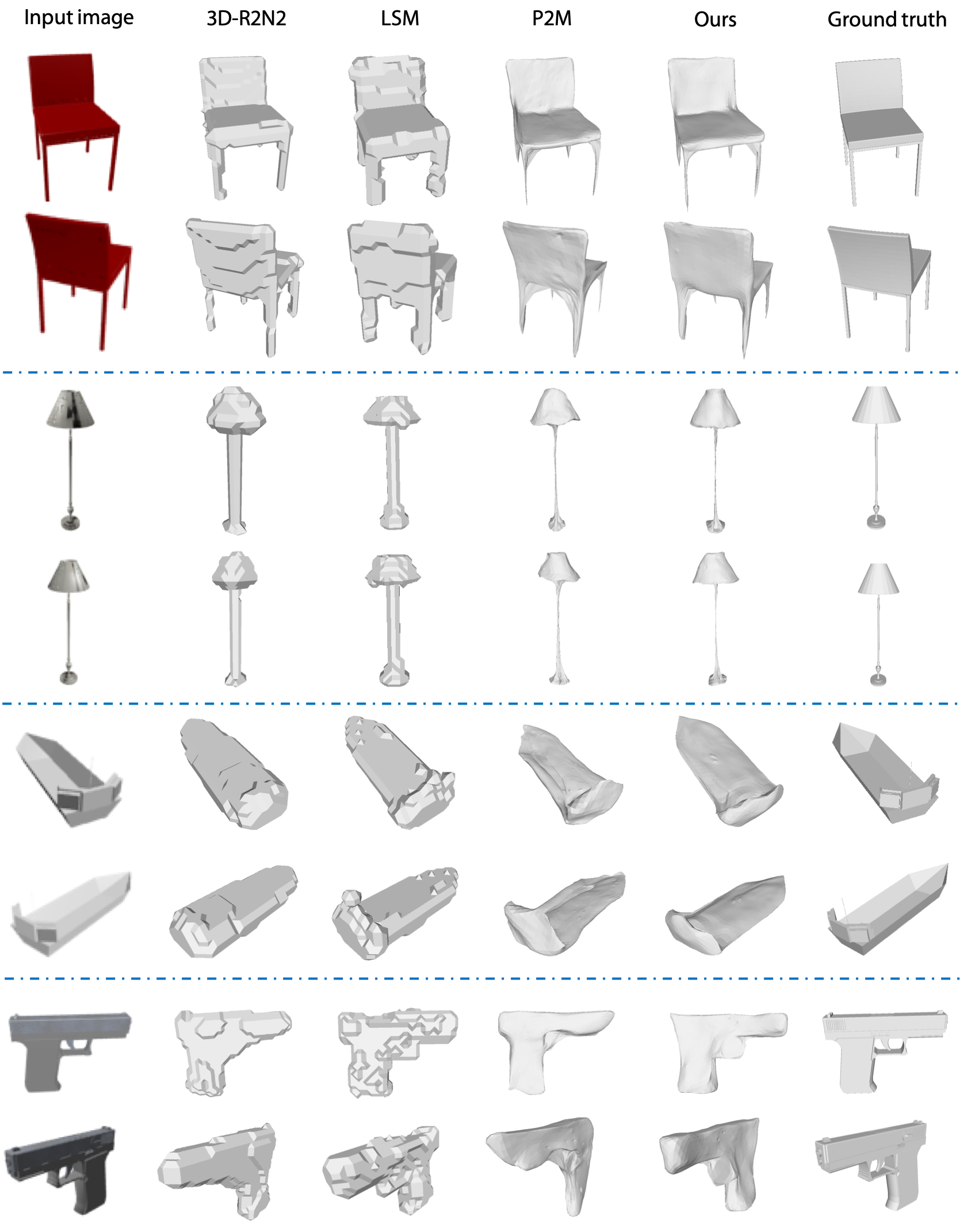}
	\caption{\textbf{More Qualitative Results.} From top to bottom, we show for each example: two camera views, results of 3DR2N2, LSM, Pixel2Mesh, ours, and the ground truth.}
	\label{fig:multiview2}
\end{figure*}

\begin{figure*}[h!]
	\centering
	\includegraphics[width=\textwidth, 
                     height=\textheight, 
                     keepaspectratio]{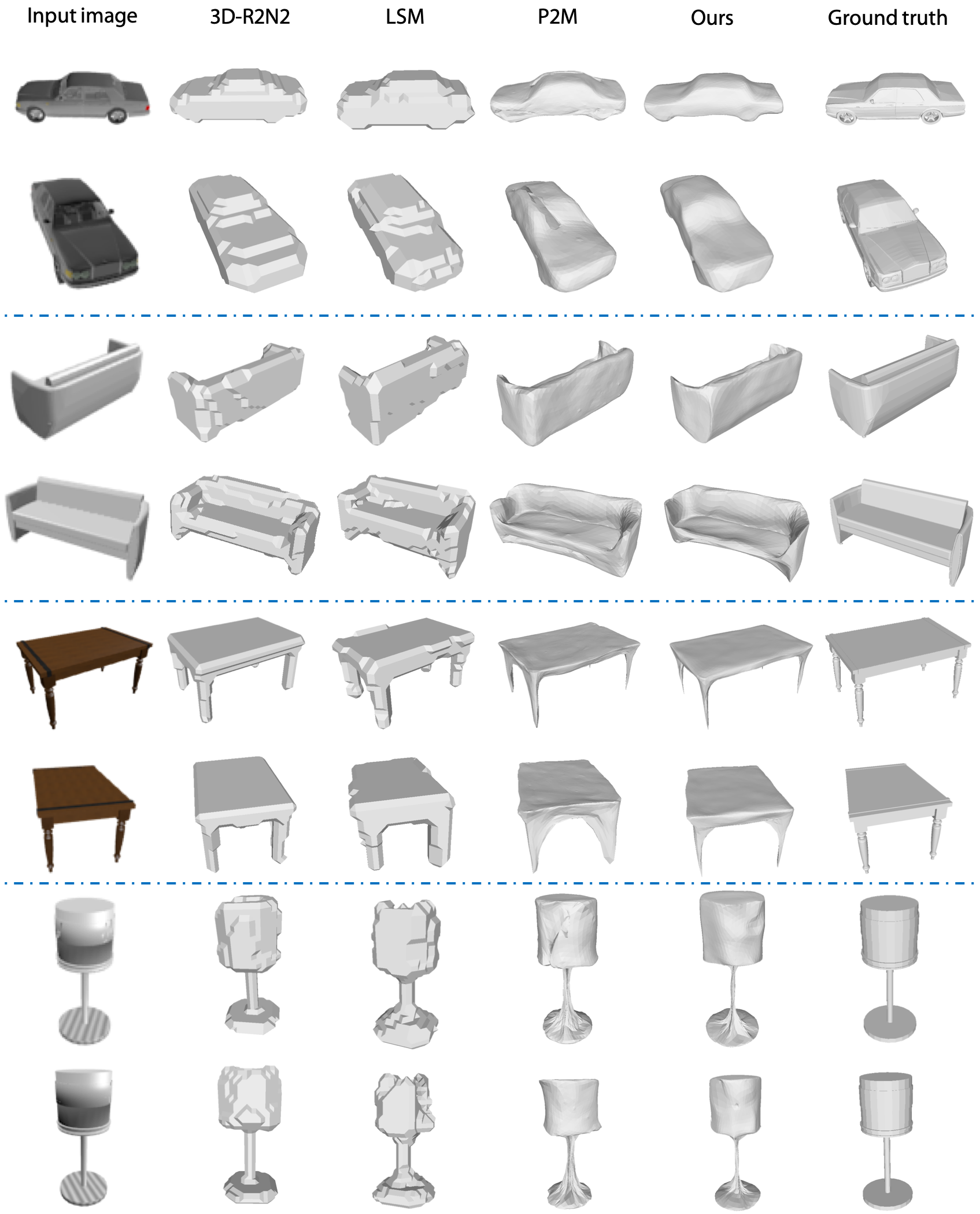}
	\caption{\textbf{More Qualitative Results.} From top to bottom, we show for each example: two camera views, results of 3DR2N2, LSM, Pixel2Mesh, ours, and the ground truth.}
	\label{fig:multiview3}
\end{figure*}

In the end, we show more qualitative results in Fig. \ref{fig:multiview}, Fig. \ref{fig:multiview2}, and Fig. \ref{fig:multiview3}.
For each example, we show the input image and the results from 3D-R2N2 \cite{choy20163d}, LSM \cite{kar2017learning}, P2M \cite{wang2018pixel2mesh}, our model, and the ground truth.
In overall, our model produces accurate shapes that align well with input views and maintain good surface details. 

\section{Discussion about Self-intersection}
Some experiments results indicate Pixel2Mesh suffers from self-intersection since it was not explicitly handled. In contrast, we observed that Pixel2Mesh++ produces results with less intersection even though we did not particularly handle it either. We conjecture that this is because geometric reasoning cross checks information from multi-view, and thus the shape generation is more stable and robust. Using more stable features and larger Laplacian regular terms in training may alleviate this problem as well.

\begin{figure*}[h!]
	\centering
	\includegraphics[width=\textwidth, 
                     height=0.9\textheight, 
                     keepaspectratio]{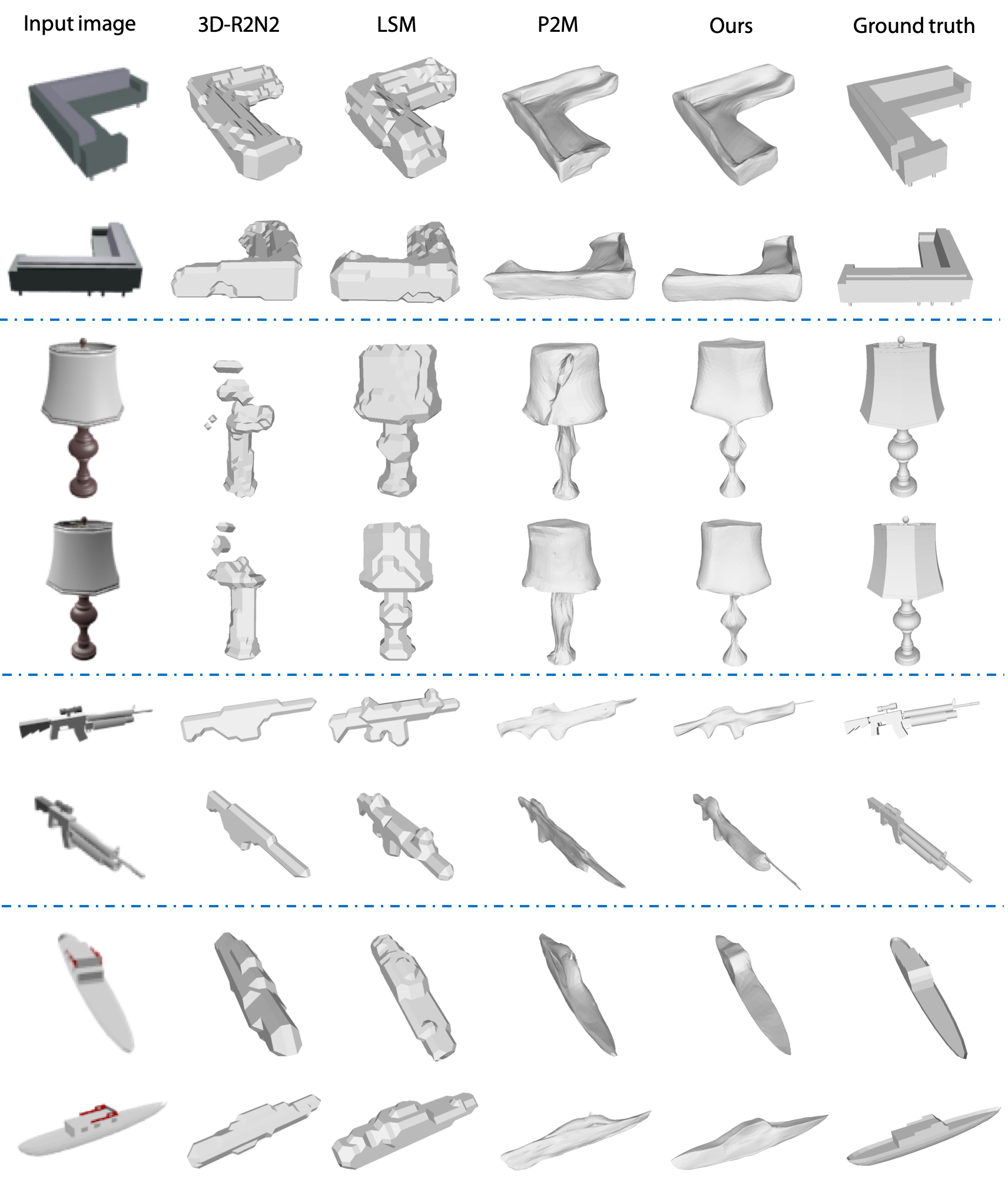}
	\caption{\textbf{More Qualitative Results.} From top to bottom, we show for each example: two camera views, results of 3DR2N2, LSM, Pixel2Mesh, ours, and the ground truth.}
	\label{fig:multiview}
\end{figure*}

\end{document}